\documentclass[10pt,journal,cspaper,compsoc]{IEEEtran}

\usepackage{url}            
\usepackage{booktabs}       
\usepackage{amsfonts}       
\usepackage{nicefrac}       
\usepackage{microtype}      
\usepackage{graphicx}
\usepackage{subfigure}
\usepackage{booktabs} 

\usepackage{url}            
\usepackage{enumitem}
\usepackage{epsfig}
\usepackage{graphicx}
\usepackage{amsmath}
\usepackage{bbm}
\PassOptionsToPackage{hyphens}{url}
\usepackage{multirow}
\usepackage{booktabs}
\usepackage{wrapfig}
\usepackage{tabularx}
\usepackage{amsmath}
\usepackage{caption}
\usepackage{float}
\usepackage{tabu}
\usepackage{amssymb}
\usepackage{wrapfig}
\usepackage{stfloats}

\usepackage[pagebackref=true,breaklinks=true,letterpaper=true,colorlinks,citecolor=black,bookmarks=false]{hyperref}

\graphicspath{ {./fig/} }

\DeclareMathOperator*{\argmin}{arg\,min}

\ifCLASSOPTIONcompsoc

  \usepackage[nocompress]{cite}

\else
  \usepackage{cite}

\fi

\newcommand{\bb}[1]{#1}
\newcommand{\rulesep}{\unskip\ \vrule width 0.01pt \ }
\newcommand{\mym}{\vspace{-0.8ex}}

\newcommand{\eqsize}{\fontsize{10pt}{10pt}\selectfont}

\newcommand\citep{\cite}
\newcommand\citet{\cite}
\renewcommand{\bb}[1]{\textbf{#1}}

\begin{document}

\title{\vspace{-1ex}Exploring Simple and Transferable Recognition-Aware Image Processing}

\author{Zhuang~Liu,
        Hung-Ju~Wang,
        Tinghui~Zhou,
        Zhiqiang~Shen,\\
        Bingyi~Kang,
        Evan~Shelhamer,
        Trevor~Darrell
\IEEEcompsocitemizethanks{
\vspace{-1ex}
\IEEEcompsocthanksitem Z. Liu, H. Wang, T. Zhou, T. Darrell are with the Department of EECS, University of California, Berkeley. 
\IEEEcompsocthanksitem Z. Shen is with the Department of ECE, Carnegie Mellon University.
\IEEEcompsocthanksitem B. Kang is with the Department of ECE, National University of Singapore.
\IEEEcompsocthanksitem E. Shelhamer is now at DeepMind. This work was done when 
he was at Adobe Research.}
}

\markboth{IEEE TRANSACTIONS ON PATTERN ANALYSIS AND MACHINE INTELLIGENCE}%
{Shell \MakeLowercase{\textit{et al.}}: Bare Demo of IEEEtran.cls for Computer Society Journals}

\IEEEtitleabstractindextext{%
\begin{abstract}
Recent progress in image recognition has stimulated the deployment of vision systems at an unprecedented scale. As a result, visual data are now often consumed not only by humans but also by machines. Existing image processing methods only optimize for better human perception, yet the resulting images may not be accurately recognized by machines. This can be undesirable, e.g., the images can be improperly handled by search engines or recommendation systems. In this work, we examine simple approaches to improve machine recognition of processed images: optimizing the recognition loss directly on the image processing network or through an intermediate input transformation model. Interestingly, the processing model's ability to enhance recognition quality can \emph{transfer} when evaluated on models of different architectures, recognized categories, tasks and training datasets. This makes the methods applicable even when we do not have the knowledge of future recognition models, e.g., when uploading processed images to the Internet. We conduct experiments on multiple image processing tasks paired with ImageNet classification and PASCAL VOC detection as recognition tasks. With these simple yet effective methods, substantial accuracy gain can be achieved with strong transferability and minimal image quality loss. Through a user study we further show that the accuracy gain can transfer to a black-box cloud model. Finally, we try to explain this transferability phenomenon by demonstrating the similarities of different models' decision boundaries. 
Code: \url{https://github.com/liuzhuang13/Transferable_RA}.  
\end{abstract}

\vspace{-1ex}
}

\maketitle

\IEEEdisplaynontitleabstractindextext
%
\IEEEpeerreviewmaketitle

\IEEEraisesectionheading{\section{Introduction}\label{sec:introduction}}

%
%
%
%

Unlike in image recognition where a deep network maps an image to a category label, a deep network used for image processing maps an input image to an output image with some desired properties. Examples include super-resolution \cite{dong2014learning}, denoising \citep{xie2012image},
deblurring \citep{eigen2013restoring}, colorization \citep{zhang2016colorful}.

The goal of such systems is to produce images of high perceptual quality to a human observer. For example, in image denoising, we aim to remove noise that is not useful to an observer and restore the image to its original ``clean'' form. Metrics like PSNR/SSIM \cite{ssim} are often used \cite{dong2014learning,srdensenet} to approximate human-perceived similarity between the processed and original images, and direct human assessment on the fidelity of the output is often considered the ``gold-standard'' \cite{srgan,zhang2018unreasonable}. Therefore, techniques have been proposed to make outputs look perceptually pleasing to humans \cite{johnson2016perceptual, srgan, pix2pix}.
 
\begin{figure}[t]
    \centering
    \includegraphics[width=0.49\textwidth]{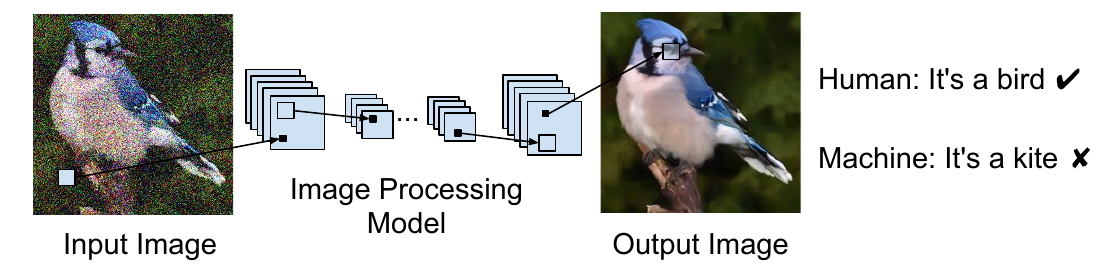}
    \caption{Image processing aims for images that look visually pleasing for human, but not those accurately recognized by machines. In this work we try to enhance output images' recognition accuracy. Zoom in for details.}
    \label{fig:fig1}
\end{figure}

 However, while looking good to humans, image processing outputs may not be accurately recognized by image recognition systems. As shown in Fig. \ref{fig:fig1}, the output image of an denoising model could easily be recognized by a human as a bird, but a recognition model classifies it as a kite.
One could specifically train a recognition model only on these output images produced by the denoising model to achieve better performance on such images, or could leverage domain adaptation approaches to adapt the recognition model to this domain, but the performance on natural images can be harmed. This retraining/adaptation scheme might also be impractical considering the significant overhead induced by catering to various image processing tasks and models. 

With the fast-growing size of image data, images are often ``viewed'' and analyzed more by machines than by humans. Nowadays, any image uploaded to the Internet is likely to be analyzed by certain vision systems.
Therefore, it is of great importance for the processed images to be recognizable \emph{not only by humans, but also by machines}. In other words, recognition systems (e.g., image classifier) should be able to accurately explain the underlying category meaning of the image content. In this way, we make them easier to search, recommended to more interested audience, and so on, as these procedures are mostly executed by machines based on their understanding of the images. 
Therefore, we argue that image processing systems should also aim at  machine recognizability.
We call this problem ``\emph{Recognition-Aware Image Processing}''.

It is also important that the enhanced recognizability is not specific to any concrete recognition model, i.e., the improvement is only achieved when the output images are evaluated on \emph{one} particular model. Instead, the improvement should ideally be \emph{transferable} when evaluated with \emph{different} downstream models/tasks, to support its usage without access to possible future recognition systems. On reason for this is what model will be used for recognizing the processed image could be out of our control, for example if we upload it to the Internet or share it on social media. We may not know what network architectures (e.g. ResNet or VGG) will be used for inference, what object categories the model recognizes (e.g. animals or scenes), or even what task will be performed (e.g. classification or detection).
Without these specifications, could it be hard for us to enhance output images' machine recognition accuracy in an unkown context?

In this work, we explore simple yet highly effective approaches to make image processing outputs more accurately recognized by downstream recognition systems, and demonstrate that these approaches generate transferable accuracy gain among different recognition architectures, categories, tasks and training datasets. 
The approaches we study add a recognition loss optimized jointly with the image processing loss. The recognition loss is computed using a fixed recognition model that is pretrained on natural images, and can be done without further supervision from class labels for training images.
It can be optimized either directly by the original image processing network, but we also have the option of resorting to an intermediate transformation network or training in an unsupervised manner, depending on different use cases.
Interestingly, we find the accuracy gain from optimizing one recognition model's loss transfers favorably among different recognition model architectures, object categories, and recognition tasks, which renders our simple solutions effective even when we do not know what the downstream recognition model is. 

We conduct extensive experiments, on multiple image processing (super-resolution, denoising,  JPEG-deblocking) and downstream recognition (classification, detection) tasks. The results demonstrate our methods can substantially boost the recognition accuracy (e.g., up to 10\%, or 20\% relative gain), with minimal loss in image quality. Results are also compared with alternative approaches in section \ref{subsec:alternatives}. We demonstrate in detail that these methods generate transferable accuracy boost, when the downstream model is either with a different architecture, recognizes different classes, performs different tasks, or even is a \emph{cloud-based, black-box} model. We conduct \emph{decision boundary analysis} and show  different models' decision boundaries exhibit high similarities, to give an explanation for this transferability phenomenon.

We would like to emphasize that in studying and analyzing these approaches, our contribution does \emph{not} lie in proposing novel network architectures, training procedures or loss functions, but in demonstrating \emph{these simple methods are surprisingly effective at making the processed images more accurately recognized}, and \emph{the improved machine recognizability can be transferred favorably to different contexts.}
The simplicity of these methods also leads to easy deployment in practice, and they could serve as strong baselines in this relatively under-explored problem. The transferability also  facilitates their practical usage in various and potentially changing deployment environments.

Our contributions can be summarized as:
\begin{itemize}
\item We propose to study the problem of enhancing the machine recognition of image processing outputs, a desired property considering the amount of images analyzed by machines nowadays.
\item We study simple yet effective methods towards this goal, suitable for different use cases. Extensive experiments are conducted on multiple image processing and recognition tasks. 

\item We show that with simple approaches, the recognition accuracy improvement is \emph{transferable} among recognition architectures, categories, tasks and datasets, a desirable behavior making the proposed methods applicable without access to downstream recognition models.
\item We provide decision boundary analysis of recognition models and show their similarities to gain a better understanding of the transferability phenomenon. 
\end{itemize}

We hope our empirical findings can encourage the community to propose new methods for improving the recognition of processed images, and further study the reason behind the intriguing transferability, which could lead to deeper understanding of neural networks.

\section{Related Work}
Image processing/enhancement problems such as super-resolution and denoising have a long history \citep{tsai1984multiframe,park2003super,rudin1992nonlinear,candes2006robust}.
Since the initial success of deep networks on these problems \citep{dong2014learning, xie2012image, wang2016d3}, a large body of works try to investigate better model architecture design and training techniques \citep{dong2016accelerating, kim2016accurate,shi2016real,Kim_2016_CVPR, lai2017deep,chen2018image}
These works focus on generating high visual quality images under conventional metrics (PSNR/SSIM) or human evaluation, without considering recognition performance on the output. 

There are also a number of works that relate image recognition with image processing. Some works \citep{zhang2016colorful, larsson2016learning, zhang2018image, sajjadi2017enhancenet, lee2019snider} use image recognition accuracy as an evaluation metric for image colorization/super-resolution/denoising, but without optimizing for it in training.
\cite{wang2016studying,haris2018task,vidalmata2019bridging,banerjee2019report,zhao2020thumbnet} investigate how to achieve more accurate recognition on low-resolution or corrupted/noisy images.
\cite{wang2019segmentation} propose a method to make denoised images more accurately segmented. \citet{liu2019classification} introduced a theoretical framework for classification-distortion-perception tradeoff and conducted experiments with simulated or toy datasets, while our work develops practical approaches for real-world datasets. Most existing works only consider one image processing task or image domain, and develop specific techniques, while our simpler approach is task-agnostic, more widely applicable, and is first to be shown transferable. Our work is related but different from those aiming for robustness of the recognition model \cite{hendrycks2019benchmarking,li2019episodic,shankar2018generalizing}, as we focus on training the processing models. Our method also shares some similarity with \citet{palacio2018deep} which tries to differentiate input signals by optimizing recognition accuracy using auto-encoders. 
\citet{bai2018finding,sicnn,liu2018disentangling,sharma2018classification,bai2018finding,liu2017image,li2018end,talebi2021learning} jointly train a processing model (e.g., dehazing, resizing, face reconstruction) together with a recognition model (e.g., object recognition or face recognition) to achieve better image processing and/or recognition quality.
Our problem setting is different from these works, in that we assume we do not have the control of the recognition model, as it might be on the cloud or to be decided in the future, thus we adapt the image processing model only. This has the advantage of making the performance gain transferable to different downstream environments, and also ensures the recognition of natural images is not harmed as the recognition model is fixed. Section \ref{subsec:alternatives} includes our comparison with training recognition model jointly.

\section{Methods}

We first formally define the problem setting of recognition-aware image processing, and then introduce the multiple approaches we examined, each suited for different use cases. We finally introduce different transferring scenarios we explored.
Our proposed methodology, although only introduced in a vision context, can be extended to other domains (e.g., speech) as well.

\subsection{Problem Setting}
\label{plain_processing}
 In a typical image processing problem, given a set of training input images $\{I^k_{in}\}$ and corresponding target images $\{I^k_{target}\}$, we aim to train a network that maps an input to its target. 
 Denoting this network as $P$ (processing), parameterized by $W_P$, our optimization objective is:
\begin{equation}
\eqsize
\label{eqn:ip_loss}
\min_{W_P} ~L_{proc} = \frac {1}{N} \sum_{k=1}^N {l_{proc} \left( P \left(I^k_{in}\right),  I^k_{target} \right)},
\end{equation}
where 
$l_{proc}$ is the loss function for each sample (e.g., $L_2$). 
The performance is typically evaluated by similarity (e.g., PSNR/SSIM) between  $I^k_{target}$ and  $P \left(I^k_{in}\right)$, or human assessment. 
In recognition-aware processing, we are interested in a recognition task, with a trained recognition model $R$ (recognition). We assume each target image $I^k_{target}$ is associated with a category label $S^k$ for the recognition task. Our goal is to train a processing model $P$ such that the recognition performance on the output images $P \left(I^k_{in}\right)$ is high, when evaluated using $R$ with the category labels $\{S^k\}$. In practice, $R$ might not be available (e.g., on the cloud), in which case we could resort to other models if the performance improvement transfers among models.

\noindent\subsection{Optimizing Recognition Loss}
\begin{figure*}[htbp]
\centering
\begin{minipage}{.49\textwidth}
 \begin{subfigure} 
 \centering
 \includegraphics[width=\textwidth]{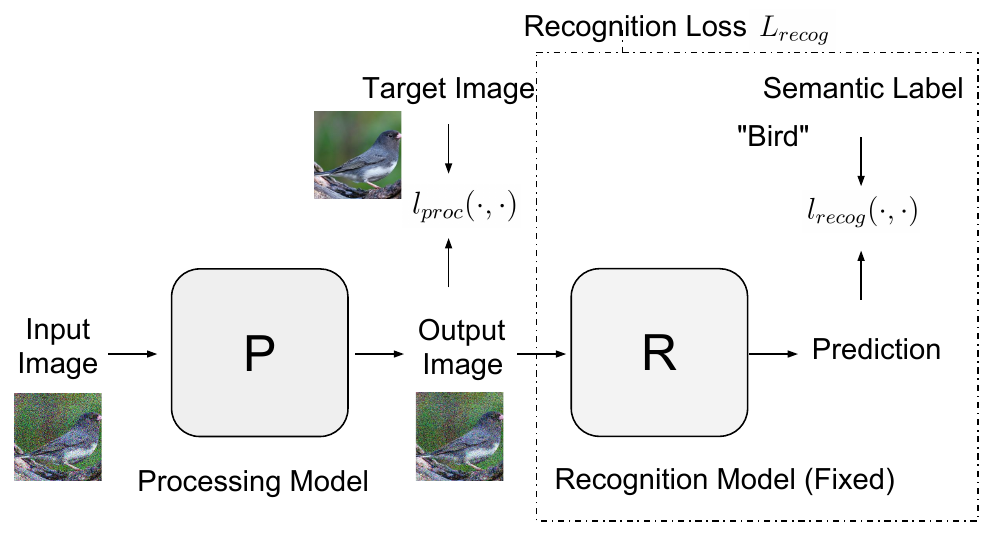}
 \label{fig:RA}
 \end{subfigure}
\end{minipage}
\rulesep
\begin{minipage}{.49\textwidth}
 \begin{subfigure} 
 \centering
 \includegraphics[width=\textwidth]{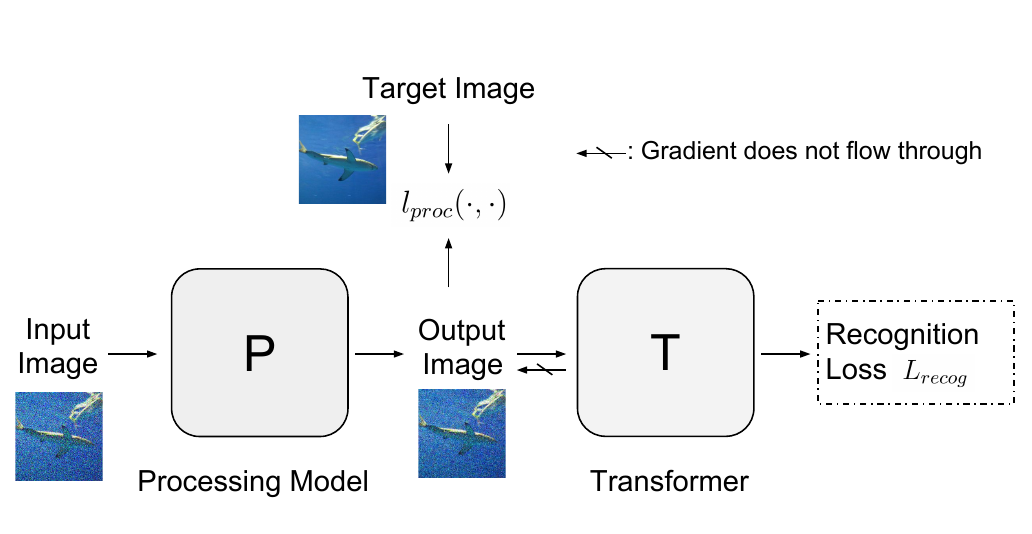}
 \label{fig:RA_T}
 \end{subfigure}
\end{minipage}
    \caption{\emph{Left}: RA (Recognition-Aware) processing. In addition to the image processing loss, we add a recognition loss using a fixed recognition model $R$, for the processing model $P$ to optimize. \emph{Right}: RA with transformer. ``Recognition Loss'' stands for the dashed box in the left figure. A Transformer $T$ is introduced between the output of $P$ and input of $R$, to optimize recognition loss. We cut the gradient from recognition loss flowing to $P$, such that $P$ only optimizes the image processing loss and the image quality is not affected.}  
    \label{fig:fig2}
\end{figure*}
Given our goal is to make the output images by $P$ more recognizable by $R$, it is natural to add a recognition loss on top of the objective of the image processing task (Eqn. \ref{eqn:ip_loss}) during training:
\begin{equation}
\label{eqn:recog_loss}
\vspace{1ex}
\eqsize
\min_{W_P} ~ L_{recog} = \frac {1}{N} \sum_{k=1}^N  {l_{recog} \left( R \left(P \left(I^k_{in}\right)\right), S^k\right)}
\vspace{1ex}
\end{equation}
$l_{recog}$ is the per-example recognition loss defined by the downstream recognition task. For example, for image classification, $l_{recog}$ could be the cross-entropy (CE) loss.
Adding the image processing loss (Eqn. \ref{eqn:ip_loss}) and recognition loss (Eqn. \ref{eqn:recog_loss}) together, our total training objective becomes 
\vspace{2ex}
\begin{equation}
\eqsize
\label{eqn:total_loss}
\min_{W_P} ~ L_{proc} + \lambda L_{recog}
\vspace{1ex}
\end{equation}
where $\lambda$ is the coefficient controlling the weights of $L_{recog}$ relative to $L_{proc}$. 
We denote this simple solution as ``RA (Recognition-Aware) processing'', which is  visualized in Fig. \ref{fig:fig2} left. Note that once the training is finished, the recognition model used as loss is not needed anymore, and during inference, we only need the processing model P, thus no overhead is introduced in deployment. A potential shortcoming of directly optimizing $L_{recog}$ is that it might deviate $P$ from optimizing the original loss $L_{proc}$, and the trained $P$ will generate images that are not as good as if we only optimize $L_{proc}$. We will show that, however, with proper choice of $\lambda$, we could substantially boost the recognition performance with nearly no sacrifice on image quality.

\subsection{Unsupervised Optimization of Recognition Loss}
The solution above requires category labels for training images, which however, may not be always available. In this case, we could regress the recognition model's output of the target image $R(I^k_{target})$, 
The recognition objective changes to
\vspace{.5ex}
\begin{equation}
\eqsize
\label{eqn:uns_recog_loss}
\min_{W_P} ~ L_{recog} = \frac {1}{N} \sum_{k=1}^N  {l_{dis} \left( R \left(P \left(I^k_{in}\right)\right), R \left(I^k_{target}\right)\right)}
\vspace{1ex}
\end{equation}

where $l_{dis}$ is a distance measure between two of $R$'s outputs (e.g., $L_2$ distance, KL divergence).
We call this approach ``unsupervised RA''. Note that it is only unsupervised for training model $P$, but not necessarily for the model $R$. The (pre)training of the model $R$ is not our concern since in our problem setting $R$ is a given trained model, and it can be trained either with or without full supervision. 
Unsupervised RA is related to the ``knowledge distillation'' paradigm \cite{kd} used for network model compression, where the output of a large model is used to guide a small model, given the same input images. Instead we use the same recognition model $R$, but guide the upstream processing model to generate desirable images. It is also related to the perceptual loss/feature loss used in \cite{johnson2016perceptual, srgan}, where the feature distance is minimized instead of output distance. We provide a comparison in Section \ref{subsec:alternatives}.

\subsection{Using an Intermediate Transformer}
Sometimes we want to prevent the added recognition loss $L_{recog}$ from causing $P$ to deviate from optimizing its original loss. We can achieve this by introducing an intermediate transformation model $T$: 
$P$'s output is first fed to the $T$, and $T$'s output serves as the input for $R$ (Fig. \ref{fig:fig2} right). $T$'s parameters $W_T$ are optimized for the recognition loss:
\begin{equation}
\eqsize
\vspace{1ex}
\label{eqn:trans_recog_loss}
\min_{W_T} ~ L_{recog} = \frac {1}{N} \sum_{k=1}^N  {l_{recog} \left( R \left( T \left( P \left(I^k_{in}\right) \right) \right), S^k\right)}
\vspace{1ex}
\end{equation}

With the help of $T$ on optimizing the recognition loss, the model $P$ can now ``focus'' on its original image processing loss $L_{proc}$. The optimization objective becomes:
\vspace{1ex}
\begin{equation}
\label{eqn:trans_total_loss}
\eqsize
\min_{W_P} ~ L_{proc} +  \min_{W_T} ~ \lambda L_{recog} 
\vspace{1ex}
\end{equation}

In Eqn. \ref{eqn:trans_total_loss}, $P$ is solely optimizing $L_{proc}$ as in the original image processing problem (Eqn. \ref{eqn:ip_loss}). $P$ is learned as if there is no recognition loss, and therefore the image processing quality of its output will not be affected. This could be implemented by ``detaching'' the gradient generated by $L_{recog}$ between the model $T$ and $P$ (Fig. \ref{fig:fig2} right). 
We term this solution as ``RA with transformer''. 
Its downside compared with directly optimization using $P$ is that there are two instances for each image (the output of model $P$ and $T$), one is ``for human'' and the other is ``for machines''. 
 Therefore, the transformer is best suited when we want to guarantee the image processing quality not affected at all, at the expense of maintaining another image. For example, in classifying images, we can have the higher-quality image presented to users for better experience and the other image passed to the backend for accurate machine classification. 
 
\subsection{Transferring Scenarios}
If using $R$ as a fixed loss can only boost the accuracy on $R$ itself, the use of the method could be limited. Sometimes we do not have the knowledge about the future downstream recognition model or even task. Thus, we explore several scenarios to see whether processing models trained with the loss of one recognition model $R_1$, can also boost the performance when evaluated using another model $R_2$. Here $R_2$ is not only possibly a different architecture but also can perform a different vision task, etc. If the improvement is transferable, then we can train our image processing model $P$ with a loss from a generic recognition model, such as those trained on ImageNet. In this case, even if in the future the output images from $P$ is tested on another model, we still have accuracy gains compared with a vanilla processing model. Specifically, we examine the following transfer scenarios:
\begin{itemize}
\item
\noindent\textbf{Transferring to a different model architecture.}  $R_1$ and $R_2$ perform the same vision task, are trained on the same datasets, and only differ in model architecture, e.g., ResNet-50 and VGG-16.

\item
\noindent\textbf{Transferring to another set of categories.} $R_1$ and $R_2$ are of the same architecture, perform the same task, but are trained to recognize disjoint subsets of categories from the same dataset. 

\item
\noindent\textbf{Transferring to another task and dataset.}
$R_1$ and $R_2$ perform different tasks, for instance image classification and object detection. In most cases, this also means $R_1$ and $R_2$ are trained on different datasets with different set of categories. They can have different architectures as well.

\item
\noindent\textbf{Transferring to a black-box model.} The model $R_2$ could be a proprietary online model that provides recognition service but does not allow users access to its structures, weights, what dataset(s) it was trained on, or even its set of output categories.
\end{itemize}

Interestingly, we find that in each of the above cases, the accuracy boost gained on $R_1$ also transfers to $R_2$. 
This makes our method effective even when we cannot access the target downstream model, where we could use another trained model as the loss function. This phenomenon also implies that the ``recognizability'' of a processed image can be more general than just the extent it fits to a specific model. More details are presented in the experiments.

\section{Experiments}
\label{sec:exp}

\newcolumntype{C}{>{\centering\arraybackslash}p{1.9em}}
\setlength{\tabcolsep}{2pt}
\renewcommand{\arraystretch}{1.2}
\begin{table*}[b]
\caption{Accuracy (\%) on ImageNet classification. R18 means ResNet-18, etc. The five models achieve 69.8, 76.2, 77.4, 74.7, 73.4 on original images. RA processing techniques substantially boost recognition accuracy. }
\mym
\centering
\small
\begin{tabular}{l|CCCCC|CCCCC|CCCCC}
\hline
\multicolumn{1}{c|}{Task} & \multicolumn{5}{c|}{Super-resolution}                & \multicolumn{5}{c|}{Denoising}                       & \multicolumn{5}{c}{JPEG-deblocking}                  \\ \hline
Classification Model      & R18      & R50      & R101     & D121     & V16      & R18      & R50      & R101     & D121     & V16      & R18      & R50      & R101     & D121     & V16      \\ \hline
No Processing             & 46.3     & 50.4     & 55.5     & 51.6     & 42.1     & 46.8     & 55.8     & 61.3     & 59.7     & 46.7     & 43.1     & 47.7     & 55.2     & 49.2     & 43.9     \\
Plain Processing          & 52.6     & 58.8     & 61.9     & 57.7     & 50.2     & 61.9     & 68.0     & 69.1     & 66.4     & 60.9     & 48.2     & 53.8     & 56.0     & 52.9     & 42.4     \\ \hline
RA Processing             & 61.8     & 67.3     & 69.6     & 66.0     & 61.9     & 65.1     & \bb{71.2} & \bb{72.7} & \bb{69.8} & \bb{66.5} & 57.7     & 63.6     & 65.8     & 62.3     & 56.7     \\
RA, Unsupervised        & 61.3     & 66.9     & 69.4     & 65.3     & 61.0     & 61.7     & 68.6     & 70.8     & 67.1     & 63.6     & 53.8     & 60.4     & 63.4     & 59.7     & 53.1     \\ 
RA w/ Transformer         & \bb{63.0} & \bb{68.2} & \bb{70.1} & \bb{66.5} & \bb{63.0} & \bb{65.2} & 70.9     & 72.3     & 69.6     & 65.9     & \bb{59.8} & \bb{65.1} & \bb{66.7} & \bb{63.9} & \bb{58.7} \\ \hline
\end{tabular}
\label{tab:cls}
\end{table*}

\begin{table*}[b]
\newcolumntype{C}{>{\centering\arraybackslash}p{1.9em}}
\caption{mAP on VOC object detection. The four models achieve 74.2, 76.8, 77.9, 72.2 on original images.}
\mym
\centering
\small
\begin{tabular}{l|CCCC|CCCC|CCCC}
\hline
\multicolumn{1}{c|}{Task} & \multicolumn{4}{c|}{Super-resolution}        & \multicolumn{4}{c|}{Denoising}                 & \multicolumn{4}{c}{JPEG-deblocking}          \\ \hline
Detection Model           & R18       & R50       & R101      & V16      & R18       & R50       & R101      & V16       & R18       & R50       & R101      & V16       \\ \hline
No Processing             & 67.9      & 70.3      & 72.1      & 63.6     & 51.8      & 56.5      & 61.8      & 38.9      & 49.3      & 54.5      & 64.1      & 38.4      \\
Plain Processing          & 69.2      & 70.7      & 73.3      & 64.2     & 68.9      & 72.0      & 74.7      & 65.8      & 63.7      & 66.5      & 70.4      & 60.3      \\ \hline
RA Processing             & 71.2      & \bb{74.4} & \bb{75.6} & \bb{68.1} & 70.9      & 73.7      & 75.6      & 67.6      & 67.4      & 70.4      & 72.9      & 63.9      \\
RA w/ Transformer         & \bb{71.4} & 74.2      & \bb{75.6} & 66.0     & \bb{71.0} & \bb{73.9} & \bb{75.9} & \bb{67.7} & \bb{68.5} & \bb{70.7} & \bb{73.7} & \bb{64.4} \\ \hline
\end{tabular}
\label{tab:det}
\label{tab:res1}
\end{table*}

\subsection{Experimental Details}
\label{subsec:details}
\textbf{General Setup.}
We evaluate our proposed methods on three image processing tasks: image super-resolution, denoising, and JPEG-deblocking. In those tasks, the target images are all the original images from the datasets. To obtain the input images, for super-resolution, we use a downsampling scale factor of 4$\times$; for denoising, we add Gaussian noise on the images with a standard deviation of 0.1 to obtain the noisy images; for JPEG deblocking, a quality factor of 10 is used to compress the image to JPEG format. The image processing loss used is the mean squared error (MSE, or $L_2$) loss. For the recognition tasks, we consider image classification and object detection, two common tasks in computer vision. In total, we have 6 (3 $\times$ 2) task pairs to evaluate. Training is performed with the training set and results on the validation set are reported. 

We adopt the SRResNet \citep{srgan} as the architecture of the image processing model $P$ (unless otherwise specified, e.g., in Sec. \ref{subsec:arch}), which is simple yet effective in optimizing the MSE loss. Even though SRResNet is originally designed for super-resolution, we find it also performs well on denoising and JPEG deblocking when its upscale parameter set to 1 for the same input-output sizes. For the transformer model $T$, we use the ResNet-like architecture in CycleGAN \citep{cyclegan}.
 The recognition models are ResNet, VGG and DenseNet. 
 Please refer to section \ref{subsec:alternatives} for comparison with alternative approaches.

Throughout the experiments, on both the image processing network and the transformer, we use the Adam optimizer \citep{adam} with an initial learning rate of $10^{-4}$, following the original SRResNet \citep{srgan}. Our implementation is in PyTorch \citep{pytorch}. 
The experiments are run on 1-4 NVIDIA TITAN Xp GPUs. 
The training process finishes in 2-24 hours depending on the model sizes/variants of methods/recognition tasks, and the maximum GPU memory taken is ~30GB (multi-GPU) with batch size of 20.

\vspace{1ex}
\noindent\textbf{Image Classification.}
For image classification, we evaluate our method on the large-scale ImageNet benchmark \citep{imagenet}, which can be downloaded at \url{http://image-net.org/download}. It consists of $\sim 1.2$ million training images and 50,000 validation images. We use five pre-trained image classification models, ResNet-18/50/101 \citep{resnet}, DenseNet-121 \citep{densenet} and VGG-16 \citep{vgg} with BN \citep{bn} (denoted as R18/50/101, D121, V16 in Table \ref{tab:cls}), on which the top-1 accuracy (\%) of the original validation images is 69.8, 76.2, 77.4, 74.7, and 73.4 respectively.
We train the processing models for 6 epochs on the training set, with a learning rate decay of 10$\times$ at epoch 5 and 6, and a batch size of 20. In evaluation, we feed unprocessed validation images to the image processing model, and report the accuracy of the output images evaluated on the pre-trained classification networks. For unsupervised RA, we use $L_2$ distance as the function $l_{dis}$ in Eqn. \ref{eqn:uns_recog_loss}. The hyperparameter $\lambda$ is chosen using super-resolution with the ResNet-18 recognition model, on two small subsets for training/validation from the original large training set, from a grid search from 10$^{-4}$ to 100. The $\lambda$ chosen for RA processing, RA with transformer, and unsupervised RA is 10$^{-3}$, 10$^{-2}$ and 10 respectively. 
\begin{table*}[ht]
\caption{Transfer between recognition architectures (ImageNet accuracy\%). A processing model trained with source model $R_A$ (row) as recognition loss can improve the recognition performance on target model $R_B$ (column).}
\mym
\centering
\footnotesize
\small
\begin{tabular}{c|CCCCC|CCCCC|CCCCC}
\hline
Task             & \multicolumn{5}{c|}{Super-resolution}                     & \multicolumn{5}{c|}{Denoising}                            & \multicolumn{5}{c}{JPEG-deblocking}                      \\ \hline
Train / Evaluation   & R18       & R50       & R101      & D121      & V16       & R18       & R50       & R101      & D121      & V16       & R18       & R50       & R101      & D121      & V16       \\ \hline
Plain Processing & 52.6      & 58.8      & 61.9      & 57.7      & 50.2      & 61.9      & 68.0      & 69.1      & 66.4      & 60.9      & 48.2      & 53.8      & 56.0      & 52.9      & 42.4      \\ \hline
RA w/ R18        & \bb{61.8} & 66.7      & 68.8      & 64.7      & 58.2      & \bb{65.1} & 70.6      & 71.9      & 69.1      & 63.8      & \bb{57.7} & 62.3      & 64.3      & 60.7      & 52.8      \\
RA w/ R50        & 59.3      & \bb{67.3} & 68.8      & 64.3      & 59.1      & 64.2      & \bb{71.2} & 72.2      & 69.2      & 64.7      & 55.8      & \bb{63.6} & 64.7      & 61.0      & 53.5      \\
RA w/ R101       & 58.8      & 66.0      & \bb{69.6} & 63.4      & 58.2      & 64.0      & 70.5      & \bb{72.7} & 68.9      & 64.8      & 54.9      & 61.5      & \bb{65.8} & 60.3      & 52.8      \\
RA w/ D121       & 59.0      & 65.6      & 67.8      & \bb{66.0} & 57.4      & 64.2      & 70.6      & 72.0      & \bb{69.8} & 64.3      & 54.8      & 61.8      & 64.4      & \bb{62.3} & 52.9      \\
RA w/ V16        & 57.9      & 64.8      & 67.0      & 63.0      & \bb{61.9} & 63.9      & 70.4      & 72.0      & 68.8      & \bb{66.5} & 54.5      & 60.9      & 63.1      & 59.7      & \bb{56.7} \\ \hline
\end{tabular}
\label{tab:transfer_model}
\end{table*}

\newcolumntype{C}{>{\centering\arraybackslash}p{1.9em}}
\setlength{\tabcolsep}{3pt}
\renewcommand{\arraystretch}{1.2}
\begin{table*}[ht]
\caption{Transfer between recognition architectures, evaluated on PASCAL VOC object detection (mAP). }
\label{tab:transfer_model_det}
\centering
\small
\begin{tabular}{c|cccc|cccc|cccc}
\hline
Task             & \multicolumn{4}{c|}{Super-resolution}         & \multicolumn{4}{c|}{Denoising}                & \multicolumn{4}{c}{JPEG-deblocking}          \\ \hline
Evaluation on    & R18       & R50       & R101      & V16       & R18       & R50       & R101      & V16       & R18       & R50       & R101      & V16       \\ \hline
Plain Processing & 69.2      & 70.7      & 73.3      & 64.2      & 68.9      & 72.0      & 74.7      & 65.8      & 63.7      & 66.5      & 70.4      & 60.3      \\ \hline
RA w/ R18        & \bb{71.2} & 73.8      & 75.2      & 66.9      & \bb{70.9} & \bb{74.0} & 75.5      & 67.2      & \bb{67.4} & 70.0      & 72.3      & 63.5      \\
RA w/ R50        & 70.6      & \bb{74.4} & 75.4      & 66.4      & 70.6      & 73.7      & 75.5      & 67.2      & 67.0      & \bb{70.4} & 72.4      & 63.2      \\
RA w/ R101       & 71.1      & 73.8      & \bb{75.6} & 65.8      & 70.3      & 73.6      & \bb{75.6} & 66.2      & 65.9      & 69.3      & \bb{72.9} & 61.3      \\
RA w/ V16        & 70.4      & 72.8      & 74.9      & \bb{68.1} & 69.9      & 73.4      & \bb{75.6} & \bb{67.6} & 66.1      & 69.3      & 72.1      & \bb{63.9} \\ \hline
\end{tabular}
\end{table*}

\vspace{1ex}
\noindent\textbf{Object Detection.}
For object detection, we evaluate on PASCAL VOC 2007 and 2012 detection dataset (\url{https://pjreddie.com/projects/pascal-voc-dataset-mirror/}), using Faster-RCNN \citep{ren2015faster} as the recognition model. Our implementation is based on the code from \citep{jjfaster2rcnn}. Following common practice \citep{yolo,ren2015faster,dai2016r}, we use VOC 07 and 12 trainval data as the training set, and evaluate on VOC 07 test data. The Faster-RCNN training uses the same hyperparameters in \citep{jjfaster2rcnn}. For the recognition model's backbone architecture, we evaluate ResNet-18/50/101 and VGG-16 (without BN \citep{bn}), obtaining mAP of 74.2, 76.8, 77.9, 72.2 on the test set respectively. Given those trained detectors as recognition loss functions, we train the models on the training set for 7 epochs, with a learning rate decay of 10 $\times$ at epoch 6 and 7, and a batch size of 1. We report the mean Average Precision (mAP) of processed images in the test set. As in image classification, we use $\lambda=10^{-3}$ for RA processing, and $\lambda=10^{-2}$ for RA with transformer.

\subsection{Evaluation on the Same Recognition Model} We first present results when the $R$ used for evaluation is the same as the $R$ we use as the recognition loss in training. Table \ref{tab:cls} shows our results on ImageNet classification. ImageNet-pretrained classification models ResNet-18/50/101, DenseNet-121 and VGG-16 are denoted as R18/50/101, D121, V16. 
``No Processing'' denotes the accuracy on input images (low-resolution/noisy/JPEG-compressed);
``Plain Processing'' denotes using image processing models trained without recognition loss (Eqn. \ref{eqn:ip_loss}).
We observe that plain processing can boost the accuracy over unprocessed images.
These two are considered as baselines.

From Table \ref{tab:cls}, using RA processing can significantly boost the accuracy of output images over plainly processed ones, for all image processing tasks and recognition models.  This is more prominent when the accuracy of plain processing is lower, e.g., in SR and JPEG-deblocking, where we mostly obtain $\sim$10\% accuracy gain (close to 20\% in relative terms). Even without category labels, our unsupervised RA can still in most cases outperform baseline methods, despite achieves lower accuracy than its supervised counterpart. Also in SR and JPEG-deblocking, using an intermediate transformer $T$ can bring additional improvement over RA processing.

The results for PASCAL VOC object detection, when evaluated on the same architecture, are shown in Table \ref{tab:det}.
We observe similar trend as in classification: using recognition loss can consistently improve the mAP over plain image processing by a notable margin. On super-resolution, RA processing mostly performs on par with RA with transformer, but on the other two tasks using a transformer is slightly better. The model with transformer performs better more often possibly because with this extra network in the middle, the capacity of the whole system is increased. 

For all result tables, each reported accuracy number is based on one run due to the relatively stable performance (almost all within 1\%) we noticed and the large amount of tasks combinations/architectures to be evaluated. For example, on super-resolution with ResNet-18 as $R$ on ImageNet, five runs on plain processing gives accuracies (\%): 53.9, 53.8, 54.0, 53.8, 53.9  (53.88$\pm$0.07); 
RA processing: 61.7, 61.9, 61.6, 61.8, 61.7 (61.74$\pm$0.10);
unsupervised RA: 61.2, 61.3, 61.3, 61.4, 61.1 (61.26$\pm$0.10);
RA w/ Transformer: 62.9, 63.0, 62.9, 62.9, 63.0  (62.94$\pm$0.05).

\subsection{Transfer between Recognition Architectures}
\label{subsec:t_arch}
In reality, sometimes the $R$ we want to eventually evaluate the output images on might not be available for us to use as a loss for training, e.g., it could be on the cloud, kept confidential or decided later. In this case, we could train an processing model $P$ using recognition model $R_A$ (source) that is accessible to us, and after we obtain the trained model $P$, evaluate its output images' accuracy using another unseen $R_B$ (target). 
We evaluate model architecture pairs on ImageNet in Table \ref{tab:transfer_model}, for RA Processing, where row is the source model ($R_A$), and column is the target model ($R_B$).
In Table \ref{tab:transfer_model}'s each column, training with any model $R_A$ produces substantially higher accuracy than plainly processed images on $R_B$, indicating that the improvement is transferable among recognition architectures. This phenomenon enables us to use RA processing without the knowledge of the downstream recognition architecture. 
\newcolumntype{C}{>{\centering\arraybackslash}p{2.2em}}
\setlength{\tabcolsep}{2pt}
\renewcommand{\arraystretch}{1.2}
\begin{table*}[htbp]
\caption{Transfer between recognition architectures using unsupervised RA, on ImageNet classification (accuracy \%).}
\label{tab:transfer_model_unsup}
\centering
\small
\begin{tabular}{l|ccccc|ccccc|ccccc}
\hline
\multicolumn{1}{c|}{Task}             & \multicolumn{5}{c|}{Super-resolution}                     & \multicolumn{5}{c|}{Denoising}                            & \multicolumn{5}{c}{JPEG-deblocking}                                                   \\ \hline
\multicolumn{1}{c|}{Evaluation on}    & R18       & R50       & R101      & D121      & V16       & R18       & R50       & R101      & D121      & V16       & R18       & R50       & R101      & \multicolumn{1}{l}{D121} & \multicolumn{1}{l}{V16} \\ \hline
\multicolumn{1}{c|}{Plain Processing} & 52.6      & 58.8      & 61.9      & 57.7      & 50.2      & \bb{61.9} & 68.0      & 69.1      & 66.4      & 60.9      & 48.2      & 53.8      & 56.0      & 52.9                     & 42.4                    \\ \hline
Unsup. RA w/ R18                      & \bb{61.3} & 66.3      & 68.6      & 64.5      & 57.3      & \bb{61.7}      & 67.9      & 69.7      & 66.4      & 60.5      & \bb{53.8} & 59.1      & 62.0      & 57.5                     & 50.0                    \\
Unsup. RA w/ R50                      & 58.9      & \bb{66.9} & 68.6      & 64.1      & 58.2      & 61.2      & \bb{68.6} & 70.3      & 66.6      & 61.3      & 52.8      & \bb{60.4} & 62.5      & 58.3                     & 50.3                    \\
Unsup. RA w/ R101                     & 57.8      & 64.9      & \bb{69.0} & 62.9      & 56.9      & 60.6      & 68.0      & \bb{70.7} & 66.3      & 60.7      & 52.3      & 58.7      & \bb{63.4} & 57.9                     & 49.0                    \\
Unsup. RA w/ D121                     & 58.0      & 64.7      & 67.2      & \bb{65.3} & 56.0      & 60.7      & 67.8      & 69.7      & \bb{67.1} & 60.3      & 52.2      & 59.2      & 62.2      & \bb{59.7}                & 49.9                    \\
Unsup. RA w/ V16                      & 57.7      & 64.6      & 67.3      & 63.2      & \bb{61.0} & 60.4      & 67.1      & 69.6      & 65.9      & \bb{63.6} & 52.0      & 58.4      & 61.5      & 57.4                     & \bb{53.1}               \\ \hline
\end{tabular}
\end{table*}

\newcolumntype{C}{>{\centering\arraybackslash}p{2.2em}}
\setlength{\tabcolsep}{2pt}
\renewcommand{\arraystretch}{1.2}
\begin{table*}[ht]
\caption{Transfer between architectures using RA with Transformer ($T$), on ImageNet classification (accuracy \%). 
}
\label{tab:transfer_model_T}
\centering
\small
\begin{tabular}{l|CCCCC|CCCCC|CCCCC}
\hline
Task             & \multicolumn{5}{c|}{Super-resolution}                     & \multicolumn{5}{c|}{Denoising}                            & \multicolumn{5}{c}{JPEG-deblocking}                      \\ \hline
Evaluation on    & R18       & R50       & R101      & D121      & V16       & R18       & R50       & R101      & D121      & V16       & R18       & R50       & R101      & D121      & V16       \\ \hline
Plain Processing & 52.6      & 58.8      & 61.9      & 57.7      & 50.2      & 61.9      & 68.0      & 69.1      & 66.4      & 60.9      & 48.2      & 53.8      & 56.0      & 52.9      & 42.4      \\ \hline
RA w/$T$ w/ R18        & \bb{63.0} & 59.2      & 67.0      & 63.9      & 27.0      & \bb{65.2} & 69.4      & 71.6      & 68.4      & 40.3      & \bb{59.8} & 58.7      & 62.6      & 60.3      & 19.9      \\
RA w/$T$ w/ R50        & 60.5      & \bb{68.2} & 68.9      & 65.8      & 40.4      & 63.1      & \bb{70.9} & 71.5      & 68.6      & 48.7      & 55.0      & \bb{65.1} & 63.9      & 61.9      & 31.5      \\
RA w/$T$ w/ R101       & 59.6      & 66.2      & \bb{70.1} & 65.1      & 35.6      & 62.4      & 68.8      & \bb{72.3} & 67.6      & 52.3      & 54.8      & 61.3      & \bb{66.7} & 24.8      & 60.5      \\
RA w/$T$ w/ D121       & 58.5      & 64.2      & 66.9      & \bb{66.5} & 27.3      & 58.0      & 66.8      & 67.3      & \bb{69.6} & 46.7      & 46.6      & 57.2      & 59.0      & \bb{63.9} & 9.0       \\
RA w/$T$ w/ V16        & 59.2      & 64.7      & 67.8      & 65.0      & \bb{63.0} & 57.6      & 64.0      & 67.1      & 55.7      & \bb{63.1} & 56.1      & 61.2      & 63.4      & 58.7      & \bb{60.1} \\ \hline
\end{tabular}
\end{table*}

We provide the model transferability results of RA processing on object detection in Table \ref{tab:transfer_model_det}. Rows indicate the models trained as recognition loss and columns indicate the evaluation models. We see similar trend as in classification (Table \ref{tab:cls}): using other architectures as loss can also improve recognition performance over plain processing; the loss model that achieves the highest performance is mostly the model itself, as can be seen from the fact that most boldface numbers are on the diagonals.

In Table \ref{tab:transfer_model_unsup}, we present the results when transferring between recognition architectures, using unsupervised RA. We note that for super-resolution and JPEG-deblocking, similar trend holds as in (supervised) RA processing, as using any architecture in training will improve over plain processing. But for denoising, this is not always the case. Some models $P$ trained with unsupervised RA are slightly worse than the plain processing counterpart. A possible reason for this is the noise level in our experiments is not large enough and plain processing achieve very high accuracy already.

In Table \ref{tab:transfer_model_T}, we present the results of transferring between architectures when we use a transformer $T$. We use the processing model $P$ and transformer $T$ trained with $R_A$ together when evaluating on $R_B$. From Table \ref{tab:transfer_model_T}, in most cases improvement is still transferable but there are a few exceptions. For example, when $R_A$ is ResNet or DenseNet and when $R_B$ is VGG-16, in most cases the accuracy fall behind plain processing by a large margin. This weaker transferability is possibly caused by the fact that there is no constraint imposed by the image processing loss on $T$'s output, thus it ``overfits'' more to the specific $R$ it is trained with.

\label{subsec:t_cat}
\subsection{Transfer between Object Categories} What if the $R_A$ and $R_B$ recognize different categories of objects? We divide the 1000 classes from ImageNet into two splits, denoted as Cat (category) $A/B$, each with 500 classes, and train two 500-way classifiers (R18) on both splits, obtaining $R_A$ and $R_B$. Next, we train two image processing models $P_A/P_B$ with the $R_A/R_B$ as recognition loss, using images from Cat $A/B$ respectively. Note that neither $P$ nor $R$ has seen images/categories from the other split.
We evaluate obtained processing models on both splits in Table \ref{tab:transfer_category}.

\setlength{\tabcolsep}{3pt}
\renewcommand{\arraystretch}{1.2}
\begin{table}[h!]
\caption{Transfer between different object categories (500-way accuracy \%). 
Refer to text in Section \ref{subsec:t_cat} for details.
}
\centering
\resizebox{0.45\textwidth}{!}{%
\begin{tabular}{l|cc|cc|cc}
\hline
\multicolumn{1}{c|}{Task}                & \multicolumn{2}{c|}{Super-res} & \multicolumn{2}{c|}{Denoising} & \multicolumn{2}{c}{JPEG-deblock} \\ \hline
\multicolumn{1}{c|}{Train/Eval} & Cat $A$           & Cat $B$           & Cat $A$        & Cat$B$        & Cat $A$           & Cat $B$          \\ \hline
Cat $A$ Plain                            & 59.6              & 60.1              & 67.6           & 68.0          & 54.2              & 55.5             \\
Cat $A$ RA                               & \bb{67.2}         & \bb{66.5}         & \bb{69.7}      & \bb{69.4}     & \bb{63.0}         & \bb{62.3}        \\ \hline
Cat $B$ Plain                            & 59.6              & 60.2              & 67.0           & 67.5          & 54.7              & 56.0             \\
Cat $B$ RA                               & \bb{66.4}         & \bb{67.8}         & \bb{69.4}      & \bb{69.7}     & \bb{62.1}         & \bb{63.5}        \\ \hline
\end{tabular}
}
\label{tab:transfer_category}
\end{table}

\newcolumntype{C}{>{\centering\arraybackslash}p{1.9em}}
\setlength{\tabcolsep}{3pt}
\renewcommand{\arraystretch}{1.2}
\begin{table*}[ht]
\caption{Transfer from ImageNet classification to PASCAL VOC object detection (mAP). 
Note that rows are classification models and columns are detection models, so even the same name in row and column (e.g., ``R18'') indicates different models trained on different tasks and datasets.
}
\mym
\centering
\footnotesize
\small
\resizebox{0.70\textwidth}{!}{%
\begin{tabular}{c|CCCC|CCCC|CCCC}
\hline
Task             & \multicolumn{4}{c|}{Super-resolution}         & \multicolumn{4}{c|}{Denoising}                & \multicolumn{4}{c}{JPEG-deblocking}          \\ \hline
Train / Evaluation    & R18       & R50       & R101       & V16        & R18       & R50       & R101      & V16       & R18       & R50       & R101      & V16       \\ \hline
Plain Processing & 68.5      & 69.7      & 73.1      & 63.2      & 68.1      & 71.6      & 74.1      & 65.7      & 62.4      & 65.6      & 69.5      & 58.3      \\ \hline
RA w/ R18         & \bb{71.3} & 73.5      & \bb{75.6} & \bb{67.8} & \bb{70.6} & 73.1      & 75.5      & 64.1      & 67.7      & 70.3      & \bb{73.2} & 62.4      \\
RA w/ R50         & 70.8      & 73.2      & 74.8      & \bb{67.8} & 70.4      & 73.1      & \bb{75.8} & 66.2      & 67.8      & 70.2      & 73.1      & 62.8      \\
RA w/ R101        & 70.7      & 73.2      & 75.3      & 67.0      & 70.5      & \bb{73.5} & 75.7      & 66.9      & \bb{68.1} & 70.2      & 72.8      & 63.2      \\
RA w/ D121        & 71.2      & \bb{73.6} & 75.3      & 67.2      & 70.5      & 73.2      & 75.7      & 65.7      & \bb{68.1} & \bb{70.5} & 73.1      & 62.6      \\
RA w/ V16         & 70.4      & 72.4      & 74.6      & 67.5      & \bb{70.6} & 73.0      & 75.7      & \bb{67.7} & 67.8      & 70.3      & \bb{73.2} & \bb{63.7} \\ \hline
\end{tabular}
}

\label{tab:transfer_task}
\end{table*}

\setlength{\tabcolsep}{2.6pt}
\renewcommand{\arraystretch}{1.2}
\begin{table*}[ht]
\caption{Transfer from PASCAL VOC object detection to ImageNet classification (accuracy \%). A image processing model $P$ trained with detection model $A$ (row) as recognition loss can improve the performance on classification model $B$ (column) over plain processing.}
\label{tab:transfer_task_det}
\centering
\small
\begin{tabular}{c|ccccc|ccccc|ccccc}
\hline
Task             & \multicolumn{5}{c|}{Super-resolution}                                    & \multicolumn{5}{c|}{Denoising}                                           & \multicolumn{5}{c}{JPEG-deblocking}                                     \\ \hline
Evaluation on    & R18       & R50       & R101      & \multicolumn{1}{l}{D121} & V16       & R18       & R50       & R101      & \multicolumn{1}{l}{D121} & V16       & R18       & R50       & R101      & \multicolumn{1}{l}{D121} & V16       \\ \hline
Plain Processing & 53.0      & 58.9      & 62.0      & 57.3                     & 50.9      & 59.7      & 65.1      & 67.3      & 63.9                     & 59.2      & 48.8      & 54.6      & 56.8      & 53.1                     & 44.7      \\ \hline
RA w/ R18        & \bb{54.6} & 60.2      & 63.4      & 58.8                     & \bb{52.7} & \bb{60.8} & \bb{66.7} & \bb{68.8} & \bb{65.2}                & \bb{61.1} & 50.8      & \bb{57.2} & \bb{59.6} & 55.4                     & \bb{48.5} \\
RA w/ R50        & 54.0      & 59.7      & 63.0      & 58.7                     & 52.0      & 60.5      & 66.6      & 68.5      & 64.9                     & 60.8      & 50.7      & 56.9      & 59.2      & 55.3                     & 48.3      \\
RA w/ R101       & 54.1      & 59.8      & 63.3      & 58.7                     & 52.5      & 60.2      & 66.1      & 68.3      & 64.6                     & 60.6      & \bb{51.3} & \bb{57.2} & 59.5      & \bb{55.5}                & 48.3      \\
RA w/ V16        & 54.5      & \bb{60.4} & \bb{63.6} & \bb{59.1}                & \bb{52.7} & 60.4      & 66.6      & 68.4      & 64.7                     & 60.6      & 50.6      & 56.5      & 58.7      & 54.9                     & 47.9      \\ \hline
\end{tabular}
\end{table*}

We observe that RA still benefits the accuracy even when transferring across categories (e.g., in SR, 60.1\% to 66.5\% transferring from Cat $A$ to Cat $B$). 
The improvement is only marginally lower than directly training on the same categories (e.g., 60.2\% to 67.8\% on Cat $B$). This suggests RA processing models do not impose category-specific signals to the images, but signals that enable wider sets of classes to be better recognized.

\subsection{Transfer between Recognition Tasks and Datasets}

\setlength{\tabcolsep}{2.6pt}
\renewcommand{\arraystretch}{1.2}
\begin{table*}[ht]
\caption{Transfer from ImageNet classification to PASCAL VOC object detection (mAP), using unsupervised RA.}
\label{tab:transfer_task_unsup}
\centering
\small
\begin{tabular}{c|cccc|cccc|cccc}
\hline
                 & \multicolumn{4}{c|}{Super-resolution}         & \multicolumn{4}{c|}{Denoising}                & \multicolumn{4}{c}{JPEG-deblocking}          \\ \hline
Evaluation on    & R18       & R50       & R101      & V16       & R18       & R50       & R101      & V16       & R18       & R50       & R101      & V16       \\ \hline
Plain Processing & 68.5      & 69.7      & 73.1      & 63.2      & 68.1      & 71.6      & 74.1      & 65.7      & 62.4      & 65.6      & 69.5      & 58.3      \\ \hline
Unsup. RA w/ R18        & \bb{71.3} & \bb{73.4} & \bb{75.3} & 66.8      & \bb{69.0} & 71.3      & 74.3      & 61.1      & 65.2      & 68.1      & 71.3      & 59.8      \\
Unsup. RA w/ R50        & 70.7      & 73.3      & 75.0      & 66.6      & 68.9      & \bb{71.7} & \bb{74.4} & 63.1      & 65.4      & 68.5      & 71.2      & 60.0      \\
Unsup. RA w/ R101       & 70.7      & 73.2      & 75.0      & 66.2      & 68.9      & 71.3      & 73.9      & 63.3      & 65.2      & 67.9      & 71.1      & 59.6      \\
Unsup. RA w/ D121       & 71.0      & 73.2      & 75.1      & 66.6      & 68.7      & 70.3      & 73.0      & \bb{63.8} & \bb{65.9} & \bb{68.6} & 71.4      & \bb{61.1} \\
Unsup. RA w/ V16        & 70.3      & 72.3      & 74.3      & \bb{67.0} & 68.5      & 70.7      & 74.0      & 63.6      & \bb{65.9} & 68.2      & \bb{71.5} & \bb{61.1} \\ \hline
\end{tabular}

\end{table*}

\label{subsec:t_task}
We evaluate task transferability when task $A$ is classification and task $B$ is object detection in Table \ref{tab:transfer_task}, where rows are classification models used for RA loss and columns are detection models for evaluation. 
There is also a dataset shift, since model $P$ and $R$ are both trained on ImageNet; during evaluation, $P$ is fed with VOC images and we use a VOC-trained detection model $R$.
We observe that 
using classification loss on model $A$ (row) gives accuracy gain on model $B$ over plain processing in most cases.
Such task transferability suggests the ``machine semantics'' of the image could be a task-agnostic property. 

 We also evaluate the opposite direction, from detection to classification. The results are shown in Table \ref{tab:transfer_task_det}. Here, using RA processing can still consistently improve over plain processing for any pair of models, but we note that the improvement is not as significant as directly training using classification models as loss (Table \ref{tab:cls} and Table \ref{tab:transfer_model}).

Additionally, the results when we transfer the model $P$ trained with unsupervised RA with image classification to object detection are shown in Table \ref{tab:transfer_task_unsup}. In most cases, it improves over plain processing, but for image denoising, this is not always the case. Similar to results in Table \ref{tab:transfer_model_unsup}, this could be because the noise level is relatively low in our experiments.

\subsection{Transfer to a Black-box, Third-party Cloud Model} We compare the images generated from plain processing and RA models using the ``General'' model at \url{clarifai.com}, a company providing state-of-the-art image classification cloud services. We do not have knowledge of the model's architecture or what datasets it was trained on, except we can access the service using APIs. The model also recognizes over 11000+ concepts that are different from the 1000-class ImageNet categories. 
For this experiment, we only take the output category with the maximum probability as the prediction. We use the SR processing model trained with R18/ImageNet as the RA model. We randomly sample image indices from ImageNet validation set, and ask clarifai.com for predictions of both images generated from plain and RA processing models. 

From the results, we then randomly select 100 instances where clarifai.com gives different predictions on plain and RA images, to compose a survey for user study. For each of the 100 instances, the survey presents the user with the target  image, and both prediction labels generated from plain/RA images, in randomized left/right order. The survey asks the user to indicate in his/her opinion which label(s) describe the image to a satisfactory level. The user has the options to choose none, either or both labels. The survey and instructions can be found at \url{https://tinyurl.com/y698779q}. 10 volunteers participated in our survey. The resulting average satisfaction rates for plain and RA super-resolved images are 40.1\% and 55.3\% respectively. We achieve 15.2\% absolute gain or 37.9\% relative gain on recognition satisfication rate, indicating the strong transferability our method provides without knowledge of the black-box cloud model.

\subsection{Experiments on More Architectures}
\label{subsec:arch}

In previous sections, we use SRResNet \citep{srgan} as our processing model $P$. Here we provide more results with other more recent processing models, including SRDenseNet (SRDNet) \citep{srdensenet}, Residual Dense Network (RDN) \citep{zhang2018residual}, and Deep Back-Projection Networks (DBPN) \cite{haris2018deep}. We present results at Table \ref{tab:P_arch}, with super-resolution as the processing task, ImageNet classification as recognition task, and $R$ being ResNet-18. The general trend we observed before holds for various archiectures.

\setlength{\tabcolsep}{2.6pt}
\renewcommand{\arraystretch}{1.2}
\begin{table}[htbp]
\small
\caption{Accuracy (\%) on ResNet-18 ImageNet classification, with more processing architectures. Processing task is supre-resolution.}
\label{tab:P_arch}
\centering
\small
\begin{tabular}{c|cccc}
\hline
 $P$ Architecture & SRResNet & SRDNet & RDN & DBPN \\
\hline
No Processing & 46.3 & 46.3 & 46.3 & 46.3 \\ 
Plain Processing & 52.6 & 54.7 & 55.5 & 54.8 \\ \hline
RA Processing & 61.8 & 62.4 & 63.8 &  62.6 \\ 
RA, Unsupervised & 61.3 & 61.7 & 63.6 & 62.4\\
RA w/ Transformer & 63.0 & 63.2 & 64.1 & 63.5\\ \hline
\end{tabular}
\end{table}

\setlength{\tabcolsep}{2.6pt}
\renewcommand{\arraystretch}{1.2}
\begin{table*}[ht]
\caption{ImageNet-C results (accuracy \%) under different types of corruptions with corruption level 5.}
\label{tab:C-RA}
\centering
\resizebox{1\textwidth}{!}{%
\begin{tabular}{c|c|ccccccccccccccccc}
\hline
Type & orig & brit & contr & defoc & elast & gau\_b & gau\_n & glass & impul & jpeg & motn & pixel & shot & satr & snow & spat & speck & zoom \\
\hline
No Processing & 69.9 & 51.3 & 3.3 & 11.3 & 17.1 & 9.3 & 1.2 & 8.7 & 1.0 & 29.4 & 11.1 & 23.1 & 1.8 & 39.5 & 10.7 & 19.1 & 7.7 & 17.6 \\
Plain Processing & N/A & 59.9 & 18.3 & 25.3 & 18.9 & 21.5 & 21.8 & 20.1 & 24.1 & 43.0 & 42.4 & 50.1 & 24.9 & 54.4 & 34.5 & 60.8 & 36.6 & 17.0      \\ \hline
RA Processing & N/A & 61.4 & 30.7 & 33.8 & 35.4 & 27.0 & 32.8 & 25.3 & 35.1 & 46.1 & 48.2 & 54.0 & 35.2 & 57.1 & 43.7 & 63.0 & 45.2 & 31.9 \\ \hline
\end{tabular}
}

\end{table*}

\setlength{\tabcolsep}{2.6pt}
\renewcommand{\arraystretch}{1.2}
\begin{table*}[ht]
\caption{ImageNet-C results (accuracy \%) evaluated with ResNet-18 as the recognition model, under different levels of corruptions, with different corruption levels of type ``snow'' and ``speckle noise''.}
\label{tab:C-level}
\centering
\small
\begin{tabular}{c|ccccc|ccccc}
\hline
                 Corruption Type & \multicolumn{5}{c|}{Snow}         & \multicolumn{5}{c}{Speckle noise}   \\ \hline
Corruption Level & 1 & 2 & 3 & 4 & 5 & 1 & 2 & 3 & 4 & 5 \\
\hline
No Processing & 46.7 & 23.6 & 28.0 & 17.6 & 10.7 & 50.5 & 42.8 & 22.9 & 14.5 & 7.7 \\
Plain Processing & 57.1 & 45.1 & 46.0 & 37.1 & 34.5 & 60.3 & 57.0 & 48.4 & 43.2 & 36.6      \\ \hline
RA Processing &  60.3 & 51.7 & 51.7 & 45.7 & 43.7 & 62.7 & 60.8 & 54.2 & 50.3 & 45.2 \\ 
Unsupervised RA & 60.2 & 51.3 & 50.6 & 43.6 & 41.5 & 62.9 & 60.5 & 53.8 & 49.4 & 43.9 \\ 
RA w/ Transformer & 55.7 & 46.7 & 48.1 & 42.7 & 40.9 & 59.0 & 57.7 & 52.2 & 49.2 & 44.7 \\ \hline
\end{tabular}
\end{table*}

\setlength{\tabcolsep}{2.6pt}
\renewcommand{\arraystretch}{1.2}
\begin{table*}[ht]
\small
\caption{ImageNet-C results (accuracy \%), trained with ResNet-18 as recognition architecture and transferred to different architectures (columns), under corruption level 5 of type ``snow'' and ``speckle noise''.}
\label{tab:C-trans}
\centering
\small
\begin{tabular}{c|ccccc|ccccc}
\hline
                 Corruption Type & \multicolumn{5}{c|}{Snow}         & \multicolumn{5}{c}{Speckle noise}   \\ \hline
Evaluation on & R18 & R50 & R101 & D121 & V16 & R18 & R50 & R101 & D121 & V16 \\
\hline
No Processing & 10.7 & 16.6 & 20.9 & 21.7 & 10.5 & 7.7 & 11.7 & 14.5 & 18.6 & 7.1  \\
Plain Processing & 34.5 & 39.1 & 44.6 & 41.1 & 27.4  & 36.6 & 42.4 & 47.7 & 43.0 & 31.3  \\ \hline
RA w/ R18 & 43.7 & 47.9 & 51.7 & 47.9 & 37.4 & 45.2 & 50.3 & 53.3 & 49.1 & 39.0 \\ 
\hline
\end{tabular}

\end{table*}

\subsection{Experiments on ImageNet-C}
\label{subsec:c}

We evaluate our methods on the ImageNet-C benchmark \citep{hendrycks2019benchmarking}. It imposes 17 different types of corruptions on the ImageNet \citep{imagenet} validation set. Despite ImageNet-C benchmark is originally designed for evaluating recognition models that are robust to corruptions, 
it is a good testbed for our methods in a broader range of processing tasks. We use the corrupted image as the input image to the processing model and the original clean image as the target image. Since only corrupted images from the validation set are released, we divide it evenly for each class into two halves and train/test on its first/second half.  The recognition model used in this experiment is an ImageNet-pretrained ResNet-18.

In Table \ref{tab:C-RA}, we evaluate RA Processing on all 17 types of corruptions, with ``corruption level'' set to 5 \citep{hendrycks2019benchmarking}. We observe that RA Processing brings consistent improvement over plain processing, sometimes by an even larger margin than the tasks considered in Sec. \ref{sec:exp}.

In Table \ref{tab:C-level}, we experiment with different levels of corruptions using two corruption types: ``speckle noise'' and ``snow''. We also evaluate with our variants -- Unsupervised RA and RA with Transformer. We observe that when the corruption level is higher, our methods tend to bring more recognition accuracy gain. In this case, we note that using a Transformer could sometimes hurt the accuracy compared with plain processing. This is possibly because the insufficient training data in ImageNet-C dataset (half of validation set) caused the transformer to hurt the accuracy, since more parameters typically require more training data. In the majority of other cases, it improves slightly over RA processing.

In Table \ref{tab:C-trans}, we examine the transferability of RA Processing between recognition architectures, using the same two tasks ``speckle noise'' and ``snow'', with corruption level 5. Note the recognition loss used during training is from a ResNet-18, and we evaluate the improvement over plain processing on ResNet-50/101, DenseNet-121 and VGG-16. The improvement over plain processing is transferable among architectures.

\subsection{Experiments on Randomized Image Corruption}
\label{subsec:realworld}

\begin{table}[h]
\newcolumntype{C}{>{\centering\arraybackslash}p{1.9em}}
\setlength{\tabcolsep}{3pt}
\renewcommand{\arraystretch}{1.2}
\captionsetup{justification=centering}
\caption{\label{tab:randomDN} Randomized and compound corruption experiment (ImageNet classification accuracy \%).}
\mym
\centering
\small
\begin{tabular}{ c|c|c } 
 \hline
    &  Plain Processing  & RA Processing \\ 
  \hline
Normal Super-resolution &  52.6    &  61.8 \\ 
 \hline
Random Super-resolution &  53.1   & 60.7 \\ 
 \hline
Normal Denoising &  61.9   & 65.1 \\ 
  \hline
Random Denoising &  59.9  & 62.1 \\ 
 \hline
Normal JPEG-deblocking &  48.2    &  57.7 \\ 
 \hline
Random JPEG-deblocking &  53.6    & 59.1 \\ 
 \hline
Random Compound & 31.5  & 41.2 \\ 
\hline

\end{tabular}
\end{table}

\newcommand{\myw}{0.158\textwidth}
\newcommand{\inc}[1]{\includegraphics[width=\myw]{#1}}
\newcommand{\s}[1]{\scriptsize{#1}}
\newcommand{\f}[1]{\scriptsize{#1}}

\setlength{\tabcolsep}{1.5pt}
\renewcommand{\arraystretch}{1.1}
\begin{figure*}[htbp]
\vspace{-6.5ex} 
\centering
\begin{tabular}{cc@{\hskip .7ex}|@{\hskip .7ex}cccc}
 \f{Target Image} & \f{Input Image} & \f{Plain Processing} & \f{RA $\lambda=10^{-4}$} &
 \f{RA $\lambda=10^{-3}$} & \f{RA $\lambda=10^{-2}$} \\
 
 \inc{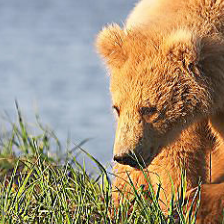} &   \inc{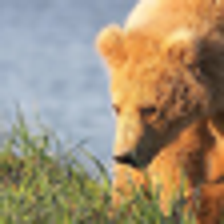} & \inc{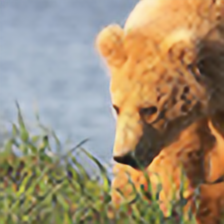} & \inc{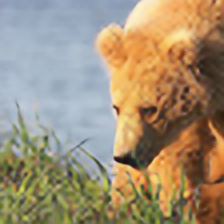} & \inc{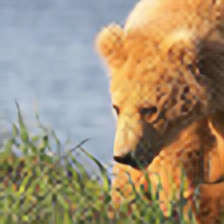} & \inc{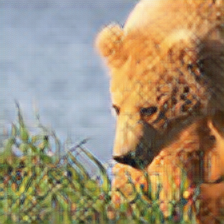} \vspace{-1ex}\\
 
\s{Label: bear}& \s{Low-resolution} & \s{19.24/0.603/lion} & \s{19.25/0.602/bear} & \s{19.20/0.600/bear} & \s{19.03/0.585/bear}\\

 \inc{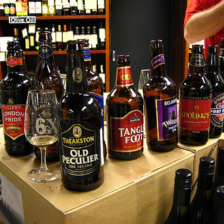} &   \inc{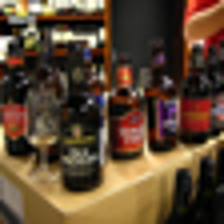} & \inc{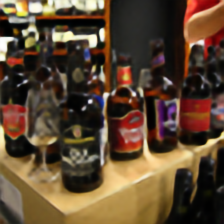} & \inc{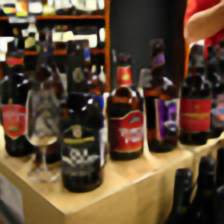} & \inc{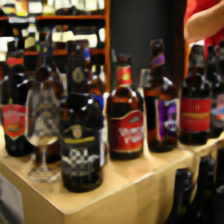} & \inc{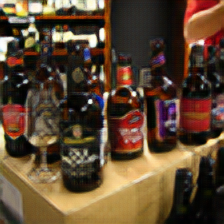} \vspace{-1ex}\\
\s{Label: beer bottle}& \s{Low-resolution} & \s{21.06/0.725/shoe shop} & \s{21.16/0.731/beer bottle} & \s{21.05/0.727/beer bottle} & \s{20.46/0.687/beer bottle}\\

 \inc{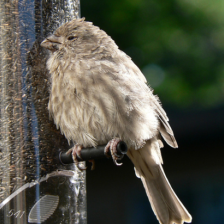} &   \inc{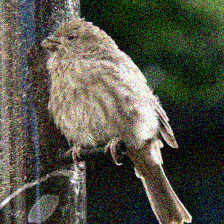} & \inc{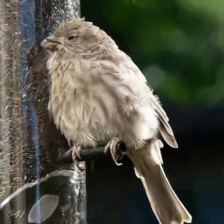} & \inc{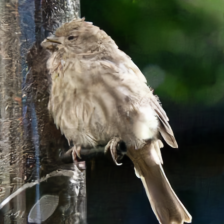} & \inc{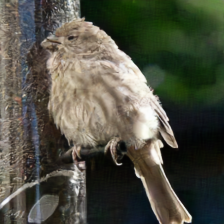} & \inc{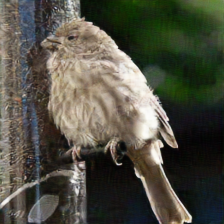} \vspace{-1ex}\\
\s{Label: finch} & \s{Noisy} & \s{30.45/0.909/kite} & \s{30.45/0.908/finch} & \s{30.18/0.899/finch} & \s{29.41/0.871/finch}  \\

\inc{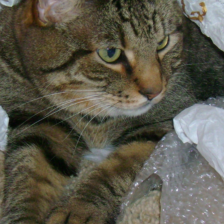} &   \inc{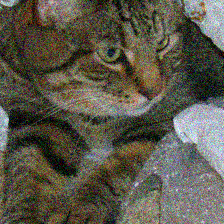} & \inc{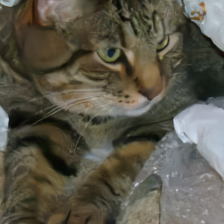} & \inc{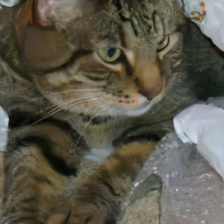} & \inc{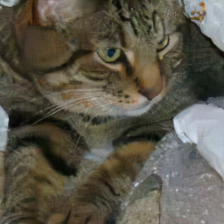} & \inc{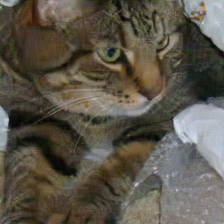}
 \vspace{-1ex}\\
\s{Label: tabby cat}& \s{Noisy} & \s{30.77/0.830/plastic bag} & \s{30.74/0.830/tabby cat} & \s{30.51/0.825/tabby cat} & \s{29.93/0.811/tabby cat}\\

 \inc{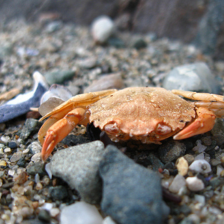} &   \inc{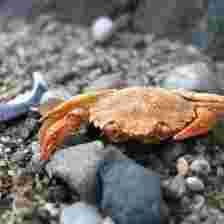} & \inc{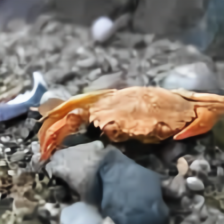} & \inc{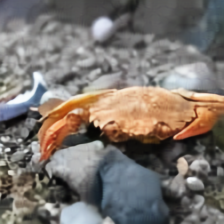} & \inc{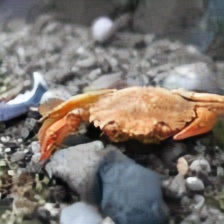} & \inc{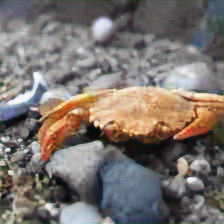} \vspace{-1ex}\\
\s{Label: crab} & \s{JPEG-compressed} & \s{27.26/0.859/goldfish} & \s{27.18/0.857/crab} & \s{26.87/0.845/crab} & \s{26.19/0.823/crab} \\

 \inc{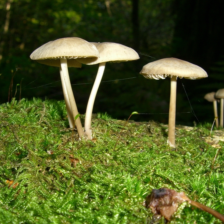} &   \inc{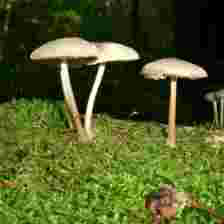} & \inc{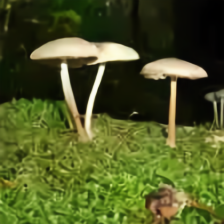} & \inc{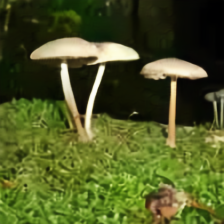} & \inc{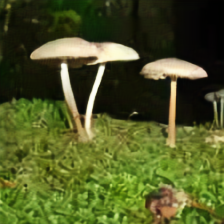} & \inc{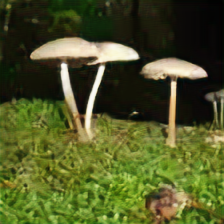} \vspace{-1ex}\\
\s{Label: mushroom}& \s{JPEG-compressed} & \s{25.78/0.746/folding chair} & \s{25.78/0.747/mushroom} & \s{21.55/0.730/mushroom} & \s{24.96/0.696/mushroom}\\
\end{tabular}
\caption{Examples where outputs of RA processing models can be correctly classified but those from plain processing models cannot. PSNR/SSIM/class prediction is shown below each output image. Slight differences between images from plain processing and RA processing models could be noticed when zoomed in.}
\label{fig:vis}
\end{figure*}
Real-world image corruptions are usually not with a fixed type or degree. To account for this fact, we develop a randomized version of image corruption for each of the image processing tasks and evaluate our methods.
For super resolution, we use a Gaussian kernel of $4\times 4$ with random standard deviations sampled from $[0.5, 1, 1.5, 2, 2.5]$; for denoising, we choose from one of the three options uniformly: 1) additive Gaussian noise with standard deviation sampled from $[0.08, 0.10, 0.12, 0.14, 0.16]$, 2) additive uniform noise with range sampled from $[0.1, 0.2, 0.3, 0.4, 0.5]$, 3) multiplicative uniform noise with range sampled from $[0.1, 0.2, 0.3, 0.4, 0.5]$; for JPEG-deblocking, the quality factor is sampled from $[5, 10, 20, 30, 40]$. We further compound these three randomized corruptions sequentially to approximate real world image distortions.

We conduct the experiments using ResNet-18 on ImageNet as the recognition model, and the results are shown in Table \ref{tab:randomDN}. In all cases, RA processing can boost the recognition accuracy. In the randomized compound experiment, the relative accuracy gain is even more significant (31.5\% $\rightarrow$ 41.2\%, a 30.8\% relative improvement).



\section{Image Processing Quality Assessment}
\label{subsec:vis}

We compare the output image quality using conventional metrics (PSNR/SSIM). When using RA with transformer, the output quality of $P$ is guaranteed unaffected, therefore here we evaluate RA processing. We use R18 as loss on ImageNet, and report results with different $\lambda$s (Eqn. \ref{eqn:total_loss}) in Table \ref{tab:experiment_quality}. $\lambda=0$ corresponds to plain processing. 

\renewcommand{\bb}[1]{\textbf{#1}}
\setlength{\tabcolsep}{3pt}
\renewcommand{\arraystretch}{1.2}
\begin{table}[ht]
\caption{PSNR/SSIM/Accuracy using different $\lambda$s.}
\centering
\resizebox{0.5\textwidth}{!}{%
\begin{tabular}{l|c|c|c}
\hline
$\lambda$ & Super-resolution & Denoising        & JPEG-deblocking   \\ \hline
0         & \bb{26.73}/\bb{0.805}/52.6 & \bb{31.24}/\bb{0.895}/61.9 & \bb{27.50}/\bb{0.825}/48.2 \\ \hline
$10^{-4}$ & 26.69/0.804/59.2 & 31.18/0.894/64.4 & \bb{27.50}/0.823/56.0 \\
$10^{-3}$ & 26.31/0.792/\bb{61.8} & 30.78/0.884/\bb{65.1} & 27.17/0.810/\bb{57.7} \\
$10^{-2}$ & 25.47/0.760/61.3 & 29.71/0.855/64.3 & 26.32/0.776/56.6 \\ \hline
\end{tabular}
}
\label{tab:experiment_quality}
\end{table}

When $\lambda=10^{-4}$, PSNR/SSIM are only marginally worse. However, the accuracy obtained is significantly higher. This suggests that the added recognition loss is not harmful when $\lambda$ is chosen properly. 
When $\lambda$ is excessively large ($10^{-2}$), image quality is hurt more, and interestingly even the recognition accuracy start to decrease, which could be due to the change of actual learning rate. A proper balance between processing and recognition loss is needed for both image quality and accuracy. We also measure the image quality using the PieAPP metric \citep{prashnani2018pieapp}, which emphasizes more on perceptual difference: on SR, when $\lambda = $ 0/10$^{-4}$/10$^{-3}$,
PieAPP (lower is better) = 1.329/1.313/1.323. Interestingly, RA processing can slightly \emph{improve} perceptual quality measured with PieAPP.
In Fig. \ref{fig:vis}, we visualize some examples where the output image is incorrectly classified with a plain processing model, but correctly recognized with RA processing. With smaller $\lambda$ ($10^{-2}$ and $10^{-3}$), the image is nearly the same as the plainly processed images. When $\lambda$ is too large ($10^{-2}$), we could see some extra textures when zooming in. 

\section{Decision Boundaries Analysis and Transferability} 
Inspired by prior works' analysis on adversarial example transferability \citep{liu2016delving, tramer2017space}, we conduct \emph{decision boundary} analysis to gain insights on RA processing's transferability. The task used is SR with ImageNet.
We restrict our analysis to a single direction at a time due to image's high dimension: given a input image $x$ and a direction $d$ (unit vector, same dimension as $x$), we analyze how the output of the recognition model $R$ changes when $x$ moves along $d$ by $\delta$ amount, i.e., when input is $x + \delta s$. We define the \emph{boundary distance} (BD) of model $R$, with respect to input $x$ and direction $s$, as the minimum amount of movement along $d$ required for $x$ to produce a different label at $R$, or more formally: $\text{BD}(R, x, d) = \argmin_{\delta >0} \{ R(x + \delta \cdot d) \neq R(x)\}$.

\begin{figure*}[t]
\centering
\vspace{2.5ex}
\includegraphics[width=1.\linewidth]{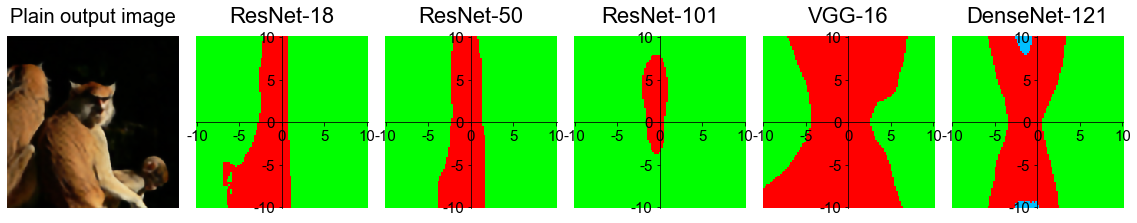}
\includegraphics[width=1.\linewidth]{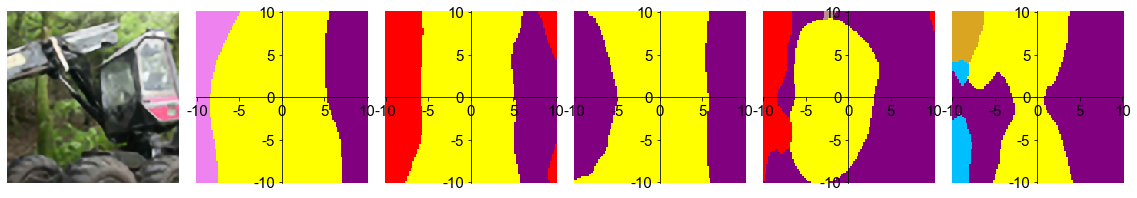}
\includegraphics[width=1.\linewidth]{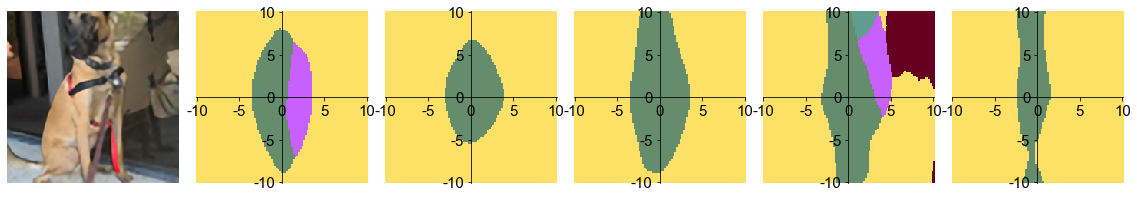}
\includegraphics[width=1.\linewidth]{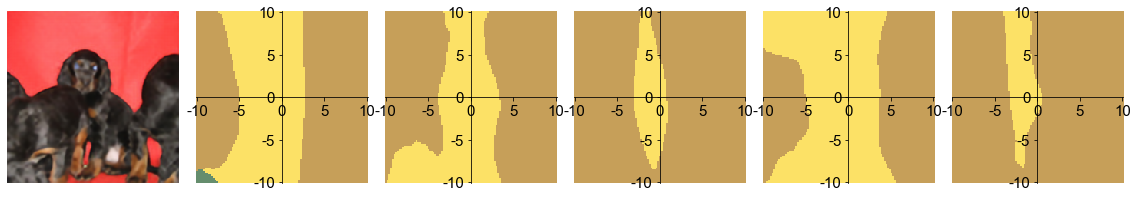}
\includegraphics[width=1.\linewidth]{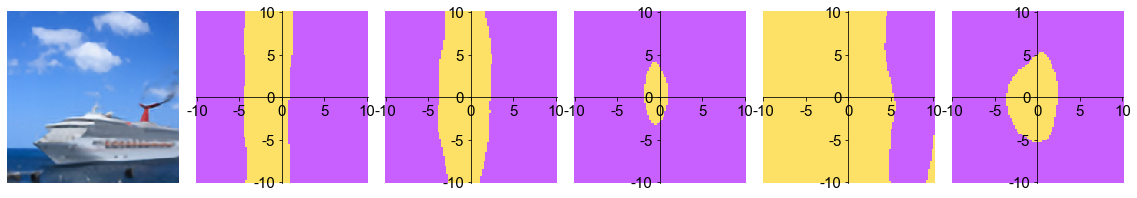}

\caption{Different models' decision boundaries are similar, especially along the RA direction (horizontal axis).}
\label{fig:boundary}
\vspace{5ex}
\end{figure*}

Consider a two-model scenario, with a source model $R_s$ and a target model $R_t$ sharing the same output categories. We define their \emph{inter-boundary distance} (IBD):  $\text{IBD}(R_s, R_t, x, d) = |\text{BD}(R_s, x, d) - \text{BD}(R_t, x, d)|$. Intuitively, if $R_s(x) = R_t(x)$ (same prediction within boundary), a small IBD between $R_s$ and $R_t$ means they have close boundaries along the $s$ direction, since $x$ does not need to move beyond one's boundary too far to reach the other's. In this case, changes made to $x$ along $d$ likely has a \emph{transferring} effect from source to target model due to their close boundaries.

We take the image $x$ to be a plain processing output, and 
consider two types of directions: 1. random direction $d_r$. 2. The direction generated by subtracting the plain processing output $x$ from RA processing output $x_s$, i.e., $(x_s-x)/||x_s-x||_2$. The RA processing model here is trained with the source model $R_s$, thus $x_s$ is specific to $R_s$.
We call this ``RA direction'' ($d_{RA}$), since it points to the RA output $x_s$ from the plain output $x$. We take all validation images such that the plain processing output $x$ generates the same wrong prediction when fed to $R_s$ and $R_t$, i.e., $R_s(x) = R_t(x) \neq \text{Ground Truth}$. For each image, we compute $\text{BD}(R_s, x, d), \text{BD}(R_t, x, d)$ and $\text{IBD}(R_s, R_t, x, d)$ with $d$ being random direction and RA direction. In this experiment we present results with $R_s$ being R18 and $R_t$ being R50, as we observe other model pairs produce similar trends.

\setlength{\tabcolsep}{6pt}
\renewcommand{\arraystretch}{1.1}
\begin{table}[ht]
\centering
\caption{Decision boundary analysis results.}
\resizebox{0.28\textwidth}{!}{%
\small
\begin{tabular}{l|c|c}
\hline
Direction             & Random & RA   \\ \hline
$\text{BD}(R_s)$      & 34.7   & 14.1 \\
$\text{BD}(R_t)$      & 34.3   & 16.5 \\
$\text{IBD}(R_s, R_t)$ & 25.2   & 10.5 \\ \hline
\end{tabular}}
\label{tab:BD}
\end{table}

The average results are shown in Table \ref{tab:BD}. We first observe that BDs are much smaller alongside the RA direction than the random direction. This indicates moving along the RA direction will change the model's wrong prediction at $x$ faster, possibly to a correct prediction. More importantly, under either random or RA direction, IBD is always smaller than source/target BDs, which indicates $R_s$ and $R_t$'s boundaries are relatively close, leading to a transfering effect. This result in RA direction can explain why RA processing can lead to transferable accuracy gains, since the RA loss brings this direction as the effect on the processing output $x$.

We further visualize decision boundaries in Fig. \ref{fig:boundary} with examples. We use R18 as source and each of the other models as target. Here we plot $d_{RA}$ as the horizontal and $d_r$ as the vertical axis. The origin represents the plain processing output $x$, and the color of point $(u, v)$ represents the predicted class of the image $x + u \cdot d_{RA} + v \cdot d_{r}$. From the plot, we can see that different models share similar decision boundaries, and also tend to change to the same prediction once we move from the origin along a direction far enough. In both examples, we do confirm that when we move towards RA direction (towards right at horizontal axis), the first color we encounter (green for top, purple for bottom) represents the image's correct label. This suggests the signal from RA loss (RA direction) can correct the wrong prediction with plain processing output ($x$ at origin), and such correction is transferable given the similar decision boundaries among models.

\section{Comparison with Alternatives}
\label{subsec:alternatives}
We analyze some alternatives to our approaches. Unless otherwise specified, experiments in this section are conducted using RA processing on super-resolution, with ResNet-18 trained on ImageNet as the recognition model, and $\lambda = 10^{-3}$ if used. 
Under this setting, we achieve 61.8\% classification accuracy on the output images.

\subsection{Training/Fine-tuning the Recognition Model} Instead of fixing the recognition model $R$, we could train/fine-tune it together with the training of image processing model $P$, to optimize the recognition loss. Many prior works \cite{sharma2018classification,bai2018finding,sicnn} do train/fine-tune the recognition model jointly with the image processing model.
We use SGD with momentum as $R$'s optimizer, and the final accuracy reaches 63.0\%. However, since we do not fix $R$, it becomes a model that specifically recognizes super-resolved images, and we found its performance on original target images drops from 69.8\% to 60.5\%. Moreover, when transferring the trained $P$ on ResNet-50, the accuracy is 62.4 \%, worse than 66.7\% when we train with a fixed ResNet-18. This suggests we lose some transferability if we do not fix the recognition model $R$.

\subsection{Training Recognition Models from Scratch}
We could first train a super-resolution model, and then train $R$ from scratch on the output images. Doing this, we achieve 66.1\% accuracy, higher than 61.8\% in RA processing. However, $R$'s accuracy on original clean images drops from 69.8\% to 66.1\%. Alternatively, we could train $R$ from scratch on interpolated low-resolution images, in which case we achieve 66.0\% on interpolated validation data but only 50.2\% on the original data. In summary, training/fine-tuning $R$ to cater the need of super-resolved or interpolated images can harm its performance on
original images, and causes additional overhead in storing models. In contrast, RA processing could boost the accuracy of output images with the performance on original images intact. 

\subsection{Training without the Image Processing Loss}
It is possible to train the processing model on the recognition loss $L_{recog}$, without even keeping the original image processing loss $L_{proc}$ (Eqn. \ref{eqn:total_loss}). This may presumably lead to better recognition performance since the model $P$ can now ``focus on'' optimizing the recognition loss. However, we found removing the original image processing loss hurts the recognition performance: the accuracy drops from 61.8\% to 60.9\%; even worse, the SSIM/PSNR metrics drop from 26.69/0.804 to 16.92/0.263, which is reasonable since the image processing loss is not optimized during training. This suggests the original image processing loss is helpful for the recognition accuracy, since it helps the corrupted image to restore to its original form.

\subsection{Perceptual/Feature Loss} Our unsupervised RA method optimizes the recognition model's output probability distance between processed and target images. This is related to the perceptual loss (also called feature loss) used in \citet{johnson2016perceptual,srgan}. Perceptual loss optimizes processed and target images' distance in VGG feature space. Note that the perceptual loss was originally proposed to improve output's quality from a human observer's perspective. To compare both methods, we follow \citet{srgan} to optimize the perceptual loss from VGG-16. We find perceptual loss yields lower accuracy than unsupervised RA (56.7\% vs. 61.0\% on the VGG-16 recognition model). This could be because using final probabilities provides more category supervision, while intermediate features improve the outputs from a perceptual perspective.

\section{Conclusion}
We investigated an important yet largely overlooked problem: enhancing the machine recognition of image processing outputs. We found that a set of simple approaches---that optimize an additional recognition loss---can significantly boost the recognition accuracy with little to no loss in image quality. Moreover, the gain in accuracy can transfer across architectures, categories, and vision tasks unseen during training, or even transfer to a black-box cloud model. This indicates that the enhanced interpretability is not specific to one particular model but generalizable to others. This makes the approaches applicable even when the future downstream recognition models are unknown. Finally we analyzed the reason of such transferability phenomenon from the perspective of decision boundary similarities between recognition models. We hope our study can encourage the community to further improve the recognition of processed images.






\ifCLASSOPTIONcaptionsoff
  \newpage
\fi


\bibliographystyle{IEEEtran}
\bibliography{ref} 

\end{document}